\documentclass{article}
\usepackage{ijcai19}

\usepackage{times}
\usepackage{soul}
\usepackage{url}
\usepackage[hidelinks]{hyperref}
\usepackage[utf8]{inputenc}
\usepackage[small]{caption}
\usepackage{graphicx}
\usepackage{amsmath}
\usepackage{booktabs}
\usepackage{algorithm}
\usepackage{algorithmic}
\newtheorem{theorem}{Theorem}
\usepackage{subfigure}

\title{A Logarithmic Barrier Method For Proximal Policy Optimization}

\author{
Cheng Zeng$^{1,2}$\footnote{Contact Author}
\And
Hongming Zhang$^3$
\affiliations
$^1$Academy of Mathematics and Systems Science, Chinese Academy of Sciences\\
$^2$University of Chinese Academy of Sciences\\
$^3$Institude of Automation, Chinese Academy of Sciences\\
\emails
zengcheng17@mails.ucas.edu.cn
}

\begin{document}

\maketitle

\begin{abstract}
Proximal policy optimization(PPO) ~\cite{Schulman2017Proximal} has been proposed as a first-order optimization method for reinforcement learning. We should notice that an exterior penalty method is used in it. Often, the minimizers of the exterior penalty functions approach feasibility only in the limits as the penalty parameter grows increasingly large. Therefore, it may result in the low level of sampling efficiency. This method, which we call proximal policy optimization with barrier method (PPO-B), keeps almost all advantageous spheres of PPO such as easy implementation and good generalization. Specifically, a new surrogate objective with interior penalty method is proposed to avoid the defect arose from exterior penalty method. Conclusions can be draw that PPO-B is able to outperform PPO in terms of sampling efficiency since PPO-B achieved clearly better performance on Atari and Mujoco environment than PPO.
\end{abstract}

\section{Introduction}

Reinforcement learning is a computational approach to understanding and automating goal-directed learning and decision making. It is distinguished from other computational approaches by its emphasis on learning by an agent from direct interaction with its environment, without requiring exemplary supervision or complete models of the environment.

The integration of reinforcement learning and neural networks has a long history ~\cite{Sutton1998Reinforcement,Bertsekas1996Neuro,Schmidhuber2015Deep}. With recent exciting achievements of deep learning ~\cite{Lecun2015Deep,Heaton2017Ian}, benefiting from big data, new algorithmic techniques and powerful computation, we have been witnessing the renaissance of reinforcement learning, especially, the combination of reinforcement learning and deep neural networks, i.e., deep reinforcement learning.

Deep reinforcement learning methods have shown tremendous success in a large variety tasks,
such as 
Atari ~\cite{Mnih2013Playing}, continuous control ~\cite{lillicrap2015continuous,Schulman2015Trust} and even Go at the human grandmaster level ~\cite{Silver2016Mastering,Silver2017Mastering}. Policy gradient methods ~\cite{Williams1992Simple} is an important family of methods in model-free reinforcement learning.

Policy gradient methods use neural network to describe the relationship between state and action, or a distribution over action. We can get the expression of the expected return under the policy $\pi$. After that, we derive the formula and write it in a form of mathematical expectation so as to get the formula of the gradient estimate under the current policy.
Then we could use Monte Carlo approach to estimate the policy gradient with the data obtained from the interaction between the current policy and the environment. After that, a gradient ascent method will be implemented on the parameters in neural networks. Once the parameters are determined finally, we actually determine an optimal policy.

There are some problems in policy gradient methods: for instance, it is in the cards to converge to the local minimum value. Sample inefficiency is also a major issue. What's more, once the optimization is done without any constraints, the parameters may update on a large scale and lead to divergence eventually. For the sake of averting this problem, TRPO creates a new constraint on the KL divergence between the two updated action distributions, and proves that under this constraint the optimal solution of our constraint problem will ensure that the expected return is increasing. However, TRPO is very complex in implement and computation. PPO algorithm is proposed to avoid this. It transform a constrain optimization to an unconstrained optimization. It is not only easy to implement, but also fairly well in experiment. In a sense, PPO is one of the best algorithm in policy gradient method.

In the original PPO, two objective functions are proposed. One is with penalty on KL divergence between two distributions, the other is a well designed pessimistic clipped surrogate objective. Experimental results showed that the PPO algorithm with the ``clipped" surrogate objective were better in more games. Although PPO with KLD penalty doesn't perform well in experiment, it can give some enlightenment to our follow-up work. From the point of view of optimization, PPO with penalty on KL divergence is actually a exterior penalty method. It constantly penalizes the larger KL divergence, forcing it to converge to the constraint region.

The exterior penalty method has some drawbacks. Though every step of iteration penalizes solutions that are not within the scope, it still can not guarantee that every update is in the feasible region, in other words, the inequality control range. When out of feasible region, the gradient will not be estimated by rule and line. Thus, in order to avoid this situation, we proposed to use the barrier function to constrain the search set of every update. Barrier functions is a family of functions which tend to infinity when it approaches the edge of constrained region, it will force every update  in the feasible region. The collected data can help us to estimate the gradient accurately, so that parameters can be updated more effectively. We conducted a complete experiment in the Atari and Mujuco environment, and achieved very competitive results relative to PPO.

The rest of this paper is organized as follows: in the first, we will elaborate the background of this paper and summarize previous work. Next, we will introduce the logarithmic barrier method. Then a new surrogate objective will be proposed to improve sampling efficiency. In the end, we will show the experience results and analysis them.

\section{Background: Policy Optimization}

\subsection{Policy Gradient}

The policy gradient methods are an important family of reinforcement learning algorithms. It assumes that the policy $\pi_{\theta}(s) $ is a map from state to an action. $\theta$ is the parameter of policy, and a fixed $\theta$ will decide a unique policy. In policy gradient method, the expected cumulative reward is written as a function of policy. That is
\begin{equation}
J^{PG}(\theta) = E_{(s_t,a_t)\sim\pi_{\theta}}[\log \pi_{\theta}(a_t|s_t)A^{\theta}]
\end{equation}
where $A^{\theta}$ is an estimator of the advantage function at timestep t. We take fixed $\theta$ to interact with the environment, and the collected data can update many $\theta$.
In this way, the gradient estimator has the form
\begin{equation}
\hat{g} = E_{(s_t,a_t)\sim\pi_{\theta}}[\nabla_{\theta} \log \pi_{\theta}(a_t|s_t)A^{\theta}]
\end{equation}
The core idea of the policy gradient algorithm is to interact with the environment using the policy, and to estimate the gradient by Monte Carlo method. In this way, we also get a new policy, and interact with the environment with the new policy. The data obtained are used to update the parameters, and this cycle improves the expected return of the policy.

\subsection{TRPO}
If the algorithm is on-policy, the data derived from  $\pi_{\theta}$ can only be used for a time. Once the $\theta$ is updated, the data collected before will be useless. This is not in line with our want. It is appealing to perform multiple steps of optimization on the loss using the same trajectory.

In this case, the more effective way is to adopt off-policy. This method needs the importance sampling. In TRPO, it gives the objective function that we need to optimize:
\begin{equation}
J^ {\theta'}(\theta)=E_{(s_t,a_t)\sim\pi_{\theta'}}[\frac{\pi_{\theta}(a_t|s_t)}{\pi_{\theta'}(a_t|s_t)}A^{\theta'}(s_t,a_t)]
\end{equation}

The $A^{\theta'} (s_t, a_t)$ is the dominant function under the parameter $\theta'$.

TRPO proves that if maximizing the upper form in every iteration, we can guarantee the expectation is monotonous. Due to the property of importance sampling, if we want to use the data more efficient, it is necessary to keep the KL divergence between old and new action distribution not very large. So what we want to deal with is a optimization problem in Equation (3) under the constraint of $KL (\pi_{\theta'} (\cdot|s), \pi_{\theta} (\cdot|s)) <\delta$.

Therefore, the constrained optimization problem proposed by TRPO is
\begin{equation}
\begin{split}
J^ {\theta'}(\theta)&=E_{(s_t,a_t)\sim\pi_{\theta'}}[\frac{\pi_{\theta}(a_t|s_t)}{\pi_{\theta'}(a_t|s_t)}A^{\theta'}(s_t,a_t)]\\
s.t. \quad& KL (\pi_{\theta'} (\cdot|s), \pi_{\theta} (\cdot|s)) <\delta
\end{split}
\end{equation}

A lot of techniques are used in TRPO to deal with this complex constrained optimization problem, such as making a linear approximation
to the objective and a quadratic approximation to the constraint, which also increase the difficulty of computation and implementation.

\subsection{PPO}

PPO method has been proposed to benefit the reliability and stability from TRPO with the goal of simpler implementation, better generalization and better empirical sample complexity. PPO algorithm puts forward two ways to improve TRPO's constrained optimization problem. The first is to use penalty method instead of constraint. Which is Shown on Equation (5). In essence, it is a exterior penalty method. For the selection of the penalty parameter, PPO proposes an adaptive way to adjust the penalty coefficient.

\begin{equation}
\begin{split}
J^{KLPEN} &= E_{(s_t,a_t)\sim\pi_{\theta'}}[\frac{\pi_{\theta}(a_t|s_t)}{\pi_{\theta'}(a_t|s_t)}A^{\theta'}(s_t,a_t)] \\
          &- \beta KL(\pi_{\theta'} (\cdot|s), \pi_{\theta} (\cdot|s))]
\end{split}
\end{equation}

The second is relatively ingenious. It replaces the original constrained problem with a ``clipped" surrogate objective, which is
\begin{equation}
\begin{split}
J^{CPI} &= E_{(s_t,a_t)\sim\pi_{\theta'}}[\frac{\pi_{\theta}(a_t|s_t)}{\pi_{\theta'}(a_t|s_t)}A^{\theta'}(s_t,a_t)] , \\ &clip(\frac{\pi_{\theta}(a_t|s_t)}{\pi_{\theta'}(a_t|s_t)},1-\epsilon,1+\epsilon)A^{\theta'}(s_t,a_t)]
\end{split}
\end{equation}

From the experimental results, PPO-clip is better.

For the first method, the exterior penalty function is used to solve the problem. However, the method of exterior penalty function will cause some problems. It can not guarantee that the two distributions are strictly in the constrain domain at each update. Therefore, this paper begins to consider the use of barrier method to ensure that parameter is strictly in the constrain region at each update. From the experimental results, our method is better than PPO-clip, and achieve state-of-art performance in policy gradient methods. We will elaborate this method in detail in the following chapters.

\section{The Logarithmic Barrier Method}

The first method in PPO are sometimes known as exterior penalty methods, because the penalty term for constraint is nonzero only when independent variables is infeasible with respect to that constraint. Often, the minimizers of the penalty functions are infeasible with respect to the original problem, and approach feasibility only in the limits as the penalty parameter grows increasingly large.

Now that exterior penalty method in PPO does not guarantee the KL divergence satisfies the constraint, a interior penalty method will be proposed here to avoid these. The interior penalty method is also called the barrier method. We introduce the concept of barrier functions by a generalized inequality-constrained optimization problem. Consider the problem
\begin{equation}
\min\limits_x f(x)\quad\mbox{subject to }c_i(x)\ge 0, \quad i\in \mathcal{I} ,
\end{equation}
the strictly feasible region is defined by
\begin{equation}
\mathcal{F}^0 \equiv \{x\in R^n | c_i(x)>0 \mbox{ for all}\ i \in \mathcal{I} \};
\end{equation}

we assume that $\mathcal{F}^0$ is nonempty for purposes of this discussion. Barrier functions for this problem have the properties that
(a)they are smooth inside $\mathcal{F}^0$;
(b)they are infinite everywhere except in $\mathcal{F}^0$;
(c)their value approaches $\infty$ as x approaches the boundary of $\mathcal{F}^0$.

The most commonly used barrier function is the logarithmic barrier function, which for the constraint set $c_i(x) \ge 0, i\in\mathcal{I}$, has the form
\begin{equation}
-\sum\limits_{i\in \mathcal{I}} \log c_i(x),
\end{equation}
where $log(\cdot)$ denotes the natural logarithm.

For the inequality-constrained optimization problem, the combined objective/barrier function is given by
\begin{equation}
L(x;\mu) = f(x) -\mu\sum\limits_{i\in\mathcal{I}}\log c_i(x)
\end{equation}
where $\mu$ is referred to here as the barrier parameter. As of now, we refer to $L(x;\mu)$ itself as the ``logarithmic barrier function'' for the Equation (7), or simply the ``log barrier function '' for short.

Consider the following problem in a single variable x:
\begin{equation}
\min x \quad \mbox{subject to}\quad x\ge 1,2-x \ge 0,
\end{equation}
for which we have
\begin{equation}
P(x;\mu) = x - \mu \log (x-1)- \mu \log(2- x).
\end{equation}
We graph this function for different values of $\mu$ in Figure 1. Naturally, for small values of $\mu$, the function $P(x;\mu)$ is close to the objective $f$ over most of the feasible set; it approaches $\infty$ only in narrow ``boundary layers.'' (In Figure 1, the curve $P(x;0.01)$ is almost indistinguishable from $f(x)$ to the resolution of our plot, though this function also approaches $\infty$ when x is very close to the endpoints 1 and 2.) Also, it is clear that as $\mu \downarrow 0$, the minimizer $x(\mu)$ of $P(x;\mu)$ is approaching the solution $x^* = 1$ of the constrained problem.

\begin{figure}[h]
\centering
\includegraphics[width=3.5in]{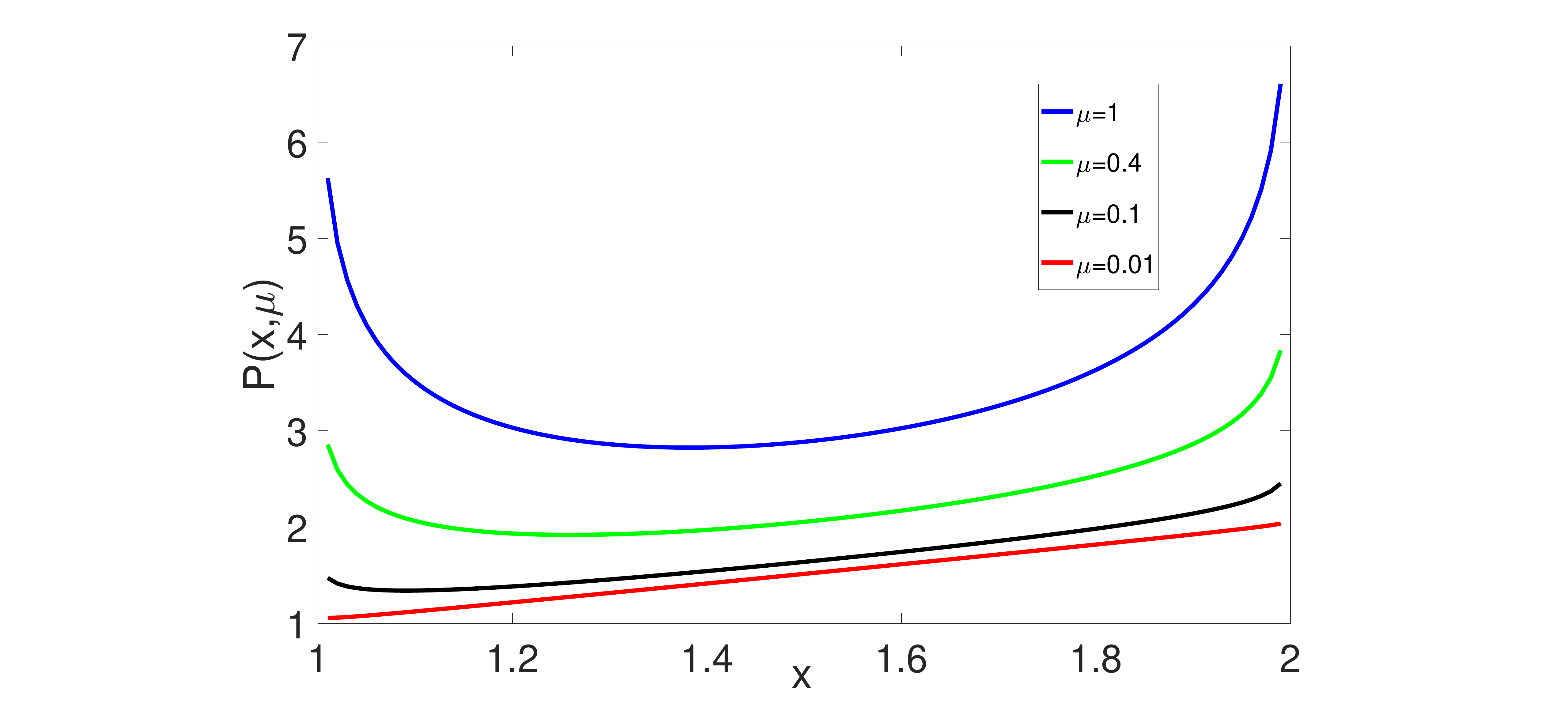}
\caption{figure of $P(x;\mu)$ for different $\mu$}
\label{fig:f1}
\end{figure}

Since the minimizer $x(\mu)$ of $P(x;\mu)$ lies in the strictly feasible set $\mathcal{F}^0$, we can in principle search for it by using the unconstrained minimization algorithms. Unfortunately, the minimizer $x(\mu)$ becomes more and more difficult to find as $\mu\downarrow 0$. So we should choose a suitable $\mu$, not as small as possible.

M. Wright ~\cite{Wright1985An} proved the effectiveness of log barrier function method in the case of convex function.
\begin{theorem}
Suppose that $f$ and $-c_i,i \in \mathcal{I}$, in Equation(7) and (8) are all convex functions, and that the strictly feasible region $\mathcal{F}^0$ defined by Equation (8) is nonempty. Assume that the solution set $\mathcal{M}$ is nonempty and bounded. Then for any $\mu >  0$, $P(x;\mu)$ is convex in $\mathcal{F}^0$ and attains a minimizer $x(\mu)$(not necessarily unique) on $\mathcal{F}^0$. Any local minimizer $x(\mu)$ is also a global minimizer of $P(x;\mu)$.
\end{theorem}

In fact, other functions can be used as barrier functions. We compare the experimental results of some barrier functions and finally choose log function as our barrier function.

\section{Barrier Method for PPO}
In this section, we will apply the logarithmic barrier method to solve constrained optimization problems of Equation (4), which will improve the Sampling Efficiency of PPO.

The essence of PPO with penalization objective is an exterior penalty method. The constrained condition is:
\begin{equation}
KL (\pi_{\theta'} (\cdot|s), \pi_{\theta} (\cdot|s)) <\delta.
\end{equation}
so we penalize the KL divergence with coefficient $\beta$, which is chosen by an adaptive way. When leaving the constraint region, the penalization is increased, forcing it to move closer to the feasible area. In the feasible area, the difference between the two is not large enough to be considered, so we can have the similar optimal solution. Unlike the exterior penalty function, we use the interior penalty method to solve this problem. The essence of exterior penalty function method is to approximate the optimal solution of the constraint problem from the outside of the feasible region, while interior penalty function contrarily. Each update will keep the constrain strictly. So this method is more suitable for solving inequality constrained optimization problems. Specifically, we use the logarithmic barrier function proposed in the previous section to solve this problem.

The barrier method is also called the interior penalty function. The minimum point of the function is a strictly feasible point, that is, a point satisfying the inequality constraint. When the minimal point sequence of the barrier function approaches the boundary of the feasible domain from the feasible domain, then the barrier function tends to infinity, so as to prevent the iteration point from falling out of the feasible region. At the same time, when we find a suitable $\mu$, the extreme value of the objective function with barrier function is closed to that of the original function, so we only need to solve an unconstrained optimization problem.

Compared with the penalty function in PPO, the objective function with barrier function is more explanatory, and according to the property of the barrier function, it can strictly ensure that the difference of distribution between two successive iterations is not too big, so that the purpose of making full use of the sample can be achieved. The experimental results in the following chapters also confirm our idea.

When applied to the problems we need to deal with, we can transform problems into
\begin{equation}
\begin{split}
J^{KLBAR}(\theta)&=E_{(s_t,a_t)\sim\pi_{\theta'}}[\frac{\pi_{\theta}(a_t|s_t)}{\pi_{\theta'}(a_t|s_t)}A^{\theta'}(s_t,a_t)]\\
&+\mu \ln [\delta - KL(\pi_{\theta}(\cdot|s_t),\pi_{\theta'}(\cdot|s_t))]
\end{split}
\end{equation}
But in practice, the distance determined by KL divergence is not a very robust distance. In practice, inspired by PPO-clip, we hope to use the distance between $\pi_{\theta} (a_t|s_t)$ and $ \pi_{\theta'} (a_t|s_t) $ to limit the difference between the two distributions. There are many ways to measure the distance between them. We have conducted several sets of controlled trials on several games, and finally chose a distance characterized by $(\sqrt{\pi_{\theta} (a_t|s_t)}-\sqrt{\pi_{\theta'} (a_t|s_t)}) ^2$. It can be easily proved that the distance is less than the Angular distance between two action distribution.

In this way, the objective function we need to optimize is changed to:
\begin{equation}
\begin{split}
J^{ADBAR} &=  E_{(s_t,a_t)\sim\pi_{\theta'}}[\frac{\pi_{\theta}(a_t|s_t)}{\pi_{\theta'}(a_t|s_t)}A^{\theta'}(s_t,a_t)] \\
&+ \mu \ln [\delta -(\sqrt{\pi_{\theta}(a_t|s_t)}-\sqrt{\pi_{\theta'}(a_t|s_t)})^2].
\end{split}
\end{equation}
We solve the whole problem in the framework of A2C ~\cite{Mnih2016Asynchronous}. It should be noted that our method will introduce two hyperparameters $\mu$ and $\delta$, but fortunately experiments show that it is sufficient to simply choose  fixed coefficients($\mu = 1$ and $\delta=0.5$) and optimize the penalized objective Equation (15) with SGD.

The pseudo code is shown on Algorithm 1. It is almost the same as the pseudo code of the PPO algorithm, except that the objective function is replaced by Equation (15).

\begin{algorithm}[h]
	\caption{POP-B, Actor-Critic Style}
	\label{pop3d}
	\begin{algorithmic}
		\STATE {\bfseries Input:} max iterations $L$ , actors $N$, epochs $K$
		\FOR{iteration=1,2 {\bfseries to} $L$}
		\FOR{ actor=1,2 {\bfseries to} $N$}
		\STATE Run policy $\pi_{\theta_{old}}$ for $T$ time steps
		\STATE Compute advantage estimations $\hat{A}_1,...,\hat{A}_T$
		\ENDFOR
		\FOR{epoch=1,2 {\bfseries to } $K$}
		\STATE Optimized loss objective wrt $\theta$ according to Equation (15) with mini-batch size $M \leq NT$, then update $\theta_{old}\leftarrow\theta$.
		\ENDFOR
		\ENDFOR
	\end{algorithmic}
\end{algorithm}

\section{Experiments}
In this section, we experimentally compare PPO-B and PPO on version-4 49 benchmark Atari game playing tasks provided by OpenAI Gym ~\cite{1606.01540} and version-2 7 benchmark control tasks provided by the robotics RL environment of PyBullet ~\cite{1802.09464}. we focus on a detailed quantitative comparison with PPO to check whether PPO-B could improve sampling efficiency and performance.

In our setting, the same policy network architecture given in ~\cite{Mnih2015Human} for Atari game playing tasks and the same network architecture given in ~\cite{Schulman2017Proximal,Duan2016Benchmarking} for benchmark control tasks are adopted in both algorithms. We also use the same training steps and make use of the same amount of game frames (40M for Atari game and 10M for Mujoco). Meanwhile we follow strictly the hyperparameters settings used in ~\cite{Schulman2017Proximal} for both PPO-B and PPO and initialize parameters using the same policy as ~\cite{Schulman2017Proximal}. The only exception is the number of actors which is set to 8 in ~\cite{Schulman2017Proximal} but equals to 16 in our experiments for both tasks. In addition to the hyperparameters used in PPO, PPO-B requires two extra hyperparameters $\delta$ and $\beta$. In our experiments, we also tested several different settings for $\mu$ and $\delta$ and chose parameters ($\mu=1$ and $\delta=0.5$) that performed best in 7 Mujoco environments.

For searching over hyperparameters for our algorithm, we used a computationally cheap benchmark proposed by ~\cite{Schulman2017Proximal} to test the algorithms on. It consists of 7 simulated robotics tasks implemented in OpenAI Gym, which use the Mujoco Environment. There are only 1 million time steps for each environment. Each algorithm was run on all 7 environments, with 3 random seeds on each. We scored each run of the algorithm by computing the average total reward of the last 100 episodes. We shifted and scaled the scores for each environment so that the random policy gave a score of 0 and the best result was set to 1, and averaged over 21 runs to produce a single score for each algorithm setting. Parameters ($\mu=1$ and $\delta=0.5$) performed best in 7 Mujoco environments, so we used the two parameters in our experiment settings.



First of all, the performance of both algorithms will be examined based on the learning curves presented in Figure 2 and Figure 3. Afterwards, we will compare the sample efficiency of PPO-B and PPO by using the performance scores summarized in Table 1 and Table 2.

\subsection{Compare With PPO on the Atari Domain}
We compared PPO-B in Atari environment with PPO algorithm. It is noteworthy that PPO used the best parameters in this experiment. The specific parameters are shown in Table 3 and Table 4. The PPO-B does not modify any of the parameters in the PPO, only adjusts the newly introduced parameters. In fact, in this case, the setting is beneficial to the PPO. We conducted experiments on 49 games on Atari and conducted three experiments in each game environment.

The final result is as follows:
\begin{table}[htbp]
\centering
\begin{tabular}{ccc}
\toprule
 & PPO & PPO-B \\
\midrule
$reward_{100}$ & 15 & $\pmb{34}$\\
$reward_{all}$ & 22 & $\pmb{27}$\\
\bottomrule
\end{tabular}
\caption{Number of games “won” by each algorithm on "Atari"}
\end{table}

We present the learning curves of the two algorithms on 49 Atari games in Figure 3. As can be clearly seen in these figures, PPO-B can outperform PPO on 34 out of the 49 games. On some game such as DemonAttack and Gopher, PPO-B performed $100\%$ better than PPO. On some games such as BreakOut and KungFuMaster, the two algorithms exhibited similar performance at the start. However PPO-B managed to achieve better performance towards the end of the learning process. On other games such as Kangaroo and Gopher, the performance differences can be witnessed shortly after learning started.

To compare PPO-B and PPO in terms of their sample efficiency, we adopt the two scoring metrics introduced in ~\cite{Schulman2017Proximal}: (1) average reward per episode over the entire training period that measures fast learning and (2) average reward per episode over last 100 episodes of training that measures final performance. As evidenced in Table 1 and Table 2, PPO-B is clearly more efficient in sampling than PPO on 34 out of 49 Atari games in the metrics of average reward per episode over last 100 episodes of training.

\subsection{Compare With PPO on the Mujoco Domain}

We also configure PPO-B and PPO in Mujoco which is a continuous environment. The two algorithms have the same parameters as those in the previous section. For each Mujoco environment, we have done three experiments on each algorithm. The experimental results are shown below. We can see that our algorithm is superior to or competitive with PPO.
\begin{table}[htbp]
\centering
\begin{tabular}{ccc}
\toprule
 & PPO &  PPO-B \\
\midrule
$reward_{100}$ & $2$ & $\pmb{5}$\\
$reward_{all}$ & $\pmb{4}$ & $3$\\
\bottomrule
\end{tabular}
\caption{Number of games “won” by each algorithm on "Mujoco"}
\label{tab1}
\end{table}

In Table 7 we present the mean of rewards of the last 100 episodes in training as a function of training time steps.
Notably, PPO-B outperforms PPO in HalfCheetah, Hopper ,InvertedDoublePendulum , InvertedPendulum and Walker2d. In the Swimmer and Reacher environment, however, we observe a different result (in Table 2) where PPO outperforms PPO-B.

\section{Conclusion}
We have proposed an improved PPO algorithm, which uses barrier method instead of exterior penalty method, and achieves good results in Atari and Mujoco environments. PPO-B makes full use of the advantages of barrier method, increasing the sampling efficiency of each actors. PPO-B also keeps the advantage of simple implementation, good generalization. It achieve better performance to the PPO algorithm, and can also give some enlightenment to the following work.
\section*{Acknowledgements}
The auther are grateful to Ruoyu Wang, Minghui Qin and others at AMSS for insightful comments.

\appendix

\bibliographystyle{named}
\bibliography{ijcai19}

\section{Hyperparameters}

\begin{table}[H]
\centering
\label{tab:booktabs}
\begin{tabular}{ccc}
\toprule
 HYPER-PARAMETER& Value  \\
\midrule
HORIZON (T) & 128 \\
ADAM STEP-SIZE & $2.5\times 10^{-4}\times\alpha $ \\
NUM EPOCHS & 3 \\
MINI-BATCH SIZE & $32\times 8$ \\
DISCOUNT($\gamma$) & 0.99\\
GAE PARAMETER($\lambda$) & 0.95 \\
NUMBER OF ACTORS & 16 \\
CLIPPING PARAMETER & $0.1\times\alpha$ \\
VF COEFF & 1\\
ENTROPY COEFF & 0.01\\
\bottomrule
\end{tabular}
\caption{PPO’s hyper-parameters for Atari game.}
\end{table}

\begin{table}[H]
\centering
\label{tab:booktabs}
\begin{tabular}{ccc}
\toprule
HYPER-PARAMETER& Value  \\
\midrule
HORIZON (T) & 128 \\
ADAM STEP-SIZE & $2.5\times 10^{-4}\times\alpha $ \\
NUM EPOCHS & 3 \\
MINI-BATCH SIZE & $32\times 8$ \\
DISCOUNT($\gamma$) & 0.99\\
GAE PARAMETER($\lambda$) & 0.95 \\
NUMBER OF ACTORS & 16 \\
VF COEFF & 1\\
ENTROPY COEFF & $0.01\times \alpha$\\
BARRIER FUNCTION PARAMETER ($\beta$) & 1 \\
BARRIER FUNCTION PARAMETER ($\delta$) & 0.5 \\
\bottomrule
\end{tabular}
\caption{PPO-B’s hyper-parameters for Atari game.}
\end{table}

\begin{table}[H]
\centering
\label{tab:booktabs}
\begin{tabular}{ccc}
\toprule
 HYPER-PARAMETER& Value  \\
\midrule
HORIZON (T) & 2048 \\
ADAM STEP-SIZE & $3\times 10^{-4} $ \\
NUM. EPOCHS & 10 \\
MINI-BATCH SIZE & $64$ \\
DISCOUNT($\gamma$) & 0.99\\
GAE PARAMETER($\lambda$) & 0.95 \\
\bottomrule
\end{tabular}
\caption{PPO’s hyper-parameters for Mujoco game.}
\end{table}

\begin{table}[H]
\centering
\label{tab:booktabs}
\begin{tabular}{ccc}
\toprule
 HYPER-PARAMETER& Value  \\
\midrule
HORIZON (T) & 2048 \\
ADAM STEP-SIZE & $3\times 10^{-4} $ \\
NUM. EPOCHS & 10 \\
MINI-BATCH SIZE & $64$ \\
DISCOUNT($\gamma$) & 0.99\\
GAE PARAMETER($\lambda$) & 0.95 \\
BARRIER FUNCTION PARAMETER ($\beta$) & 1 \\
BARRIER FUNCTION PARAMETER ($\delta$) & 0.5 \\
\bottomrule
\end{tabular}
\caption{PPO-B’s hyper-parameters for Mujoco.}
\end{table}

\section{Performance on Mujoco Games}

\begin{figure}[H]
\centering
\subfigure{
\includegraphics[width=0.3\linewidth]{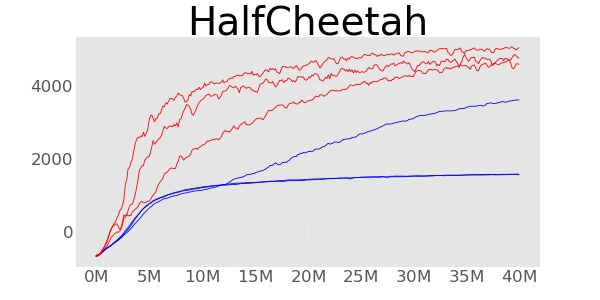}}
\hspace{0.01\linewidth}
\subfigure{
\includegraphics[width=0.3\linewidth]{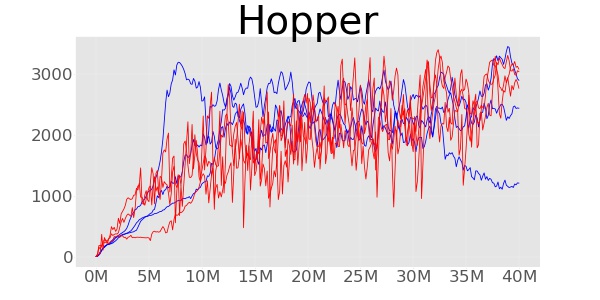}}
\hspace{0.01\linewidth}
\subfigure{
\includegraphics[width=0.3\linewidth]{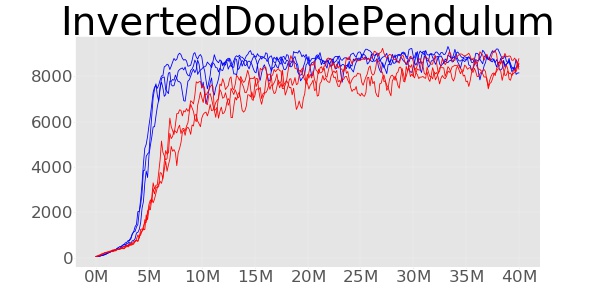}}

\subfigure{
\includegraphics[width=0.3\linewidth]{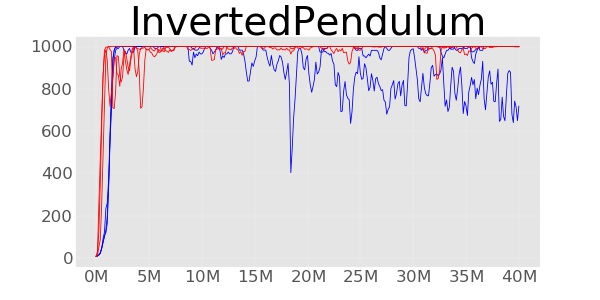}}
\hspace{0.01\linewidth}
\subfigure{
\includegraphics[width=0.3\linewidth]{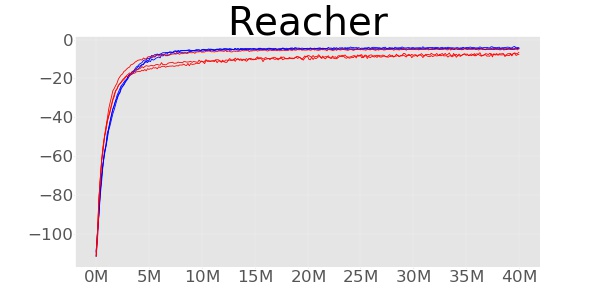}}
\hspace{0.01\linewidth}
\subfigure{
\includegraphics[width=0.3\linewidth]{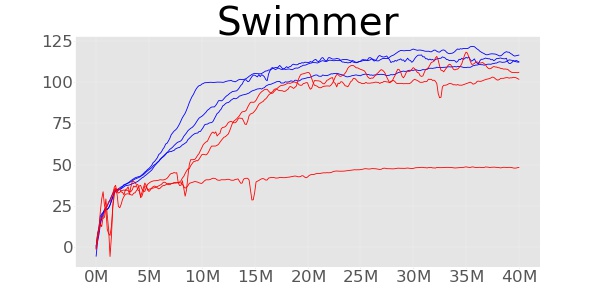}}

\subfigure{
\includegraphics[width=0.3\linewidth]{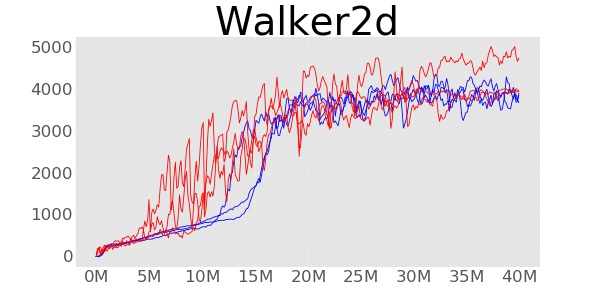}}
\hspace{0.01\linewidth}
\subfigure{
\includegraphics[width=0.3\linewidth]{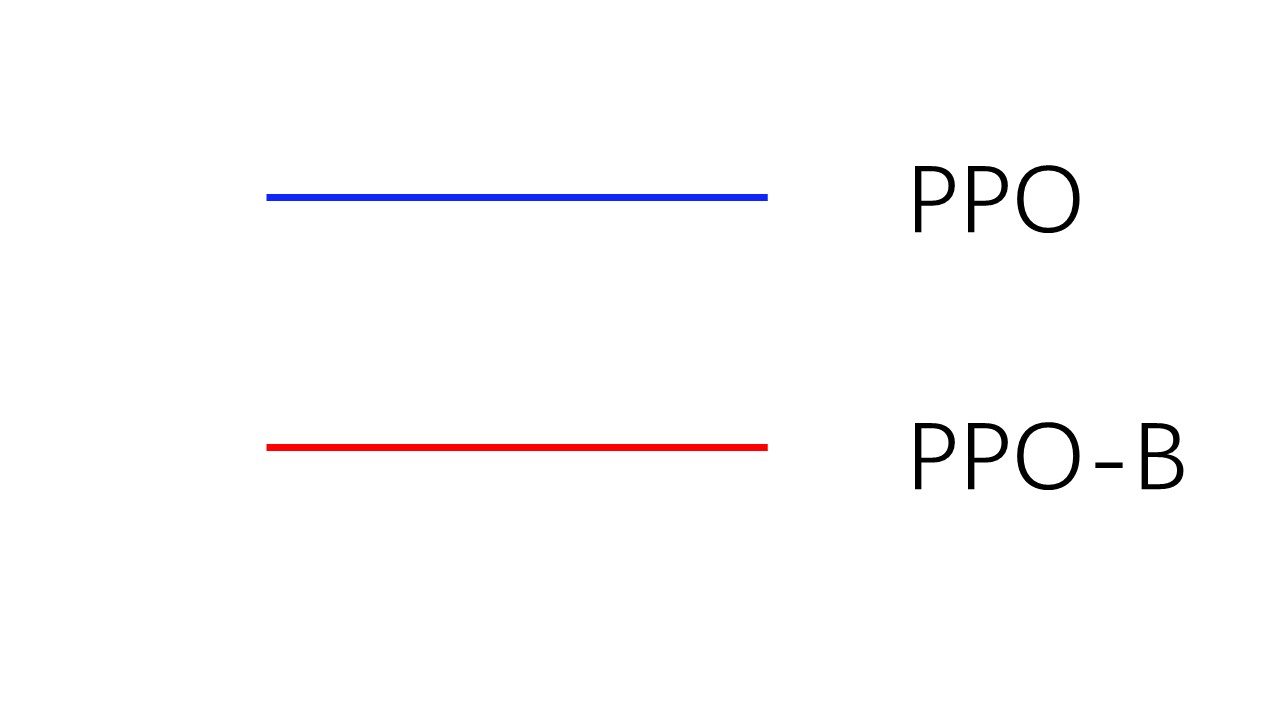}}

\caption{Comparison of several algorithms on several MuJoCo environments}
\end{figure}

\begin{table}[H]
\centering
\label{tab:booktabs}
\begin{tabular}{ccc}
\toprule
 & PPO-B & PPO  \\
\midrule
HalfCheetah & \pmb{4784.27} & 2252.16\\
Hopper & \pmb{2968.56} & 2187.83\\
InvertedDoublePendulum & \pmb{8562.62} &8377.86 \\
InvertedPendulum & \pmb{999.43} & 905.88\\
Reacher & -6.31 & \pmb{-4.51}\\
Swimmer & 85.28& \pmb{113.57}\\
Walker2d & \pmb{4201.24} & 3794.44\\

\bottomrule
\end{tabular}
\caption{Mean final scores (last 100 episodes) of PPO-B and PPO on Mujoco}
\end{table}

\section{Performance on More Atari Games}
\begin{figure}[h]
\centering
\subfigure{
\includegraphics[width=0.2\linewidth]{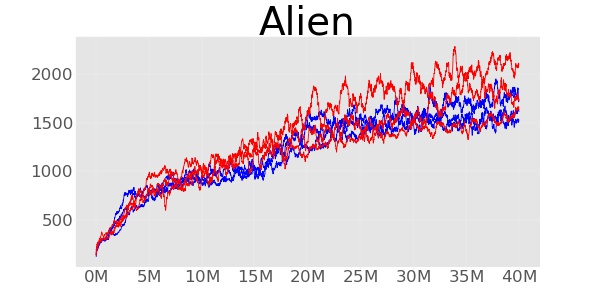}}
\hspace{0.01\linewidth}
\subfigure{
\includegraphics[width=0.2\linewidth]{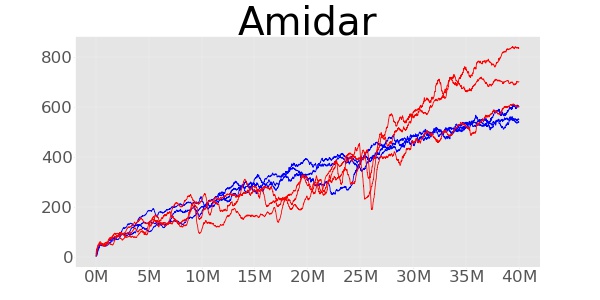}}
\hspace{0.01\linewidth}
\subfigure{
\includegraphics[width=0.2\linewidth]{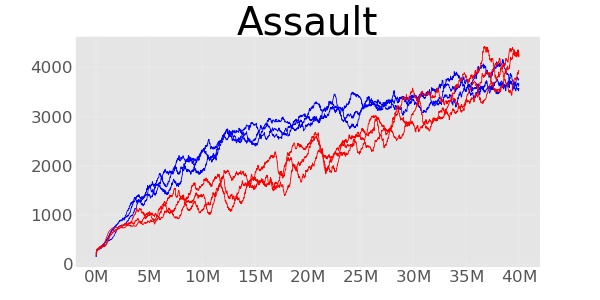}}
\hspace{0.01\linewidth}
\subfigure{
\includegraphics[width=0.2\linewidth]{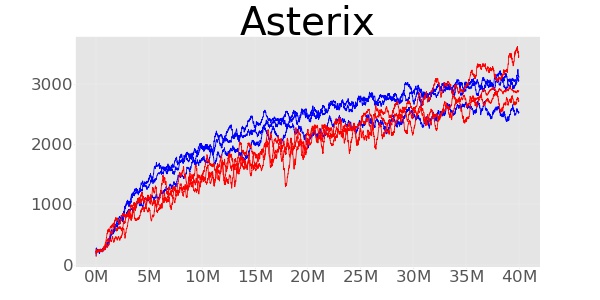}}

\subfigure{
\includegraphics[width=0.2\linewidth]{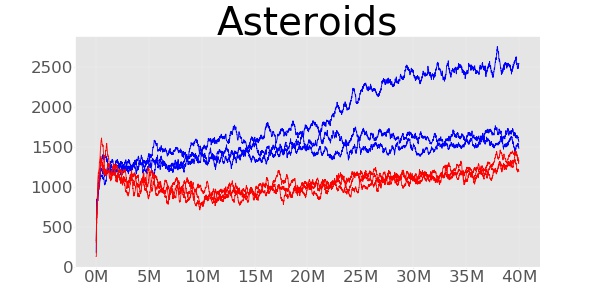}}
\hspace{0.01\linewidth}
\subfigure{
\includegraphics[width=0.2\linewidth]{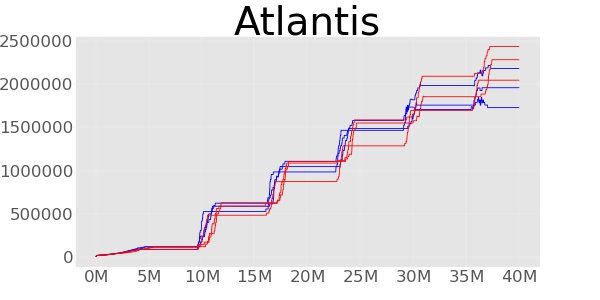}}
\hspace{0.01\linewidth}
\subfigure{
\includegraphics[width=0.2\linewidth]{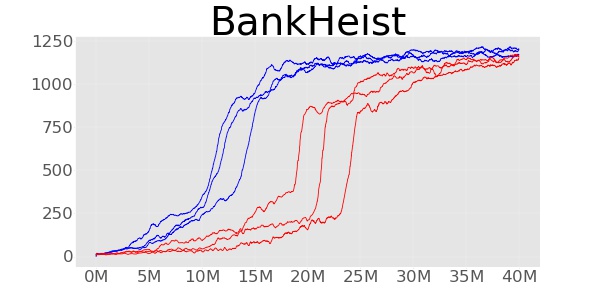}}
\hspace{0.01\linewidth}
\subfigure{
\includegraphics[width=0.2\linewidth]{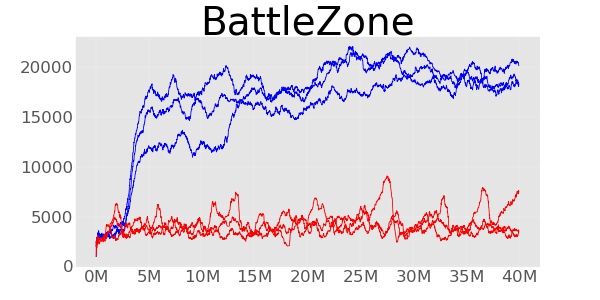}}

\subfigure{
\includegraphics[width=0.2\linewidth]{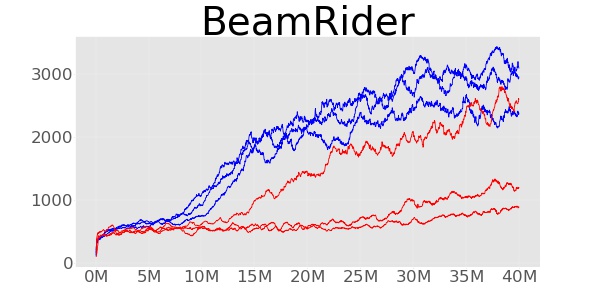}}
\hspace{0.01\linewidth}
\subfigure{
\includegraphics[width=0.2\linewidth]{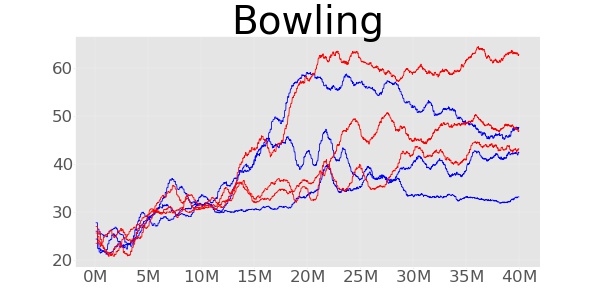}}
\hspace{0.01\linewidth}
\subfigure{
\includegraphics[width=0.2\linewidth]{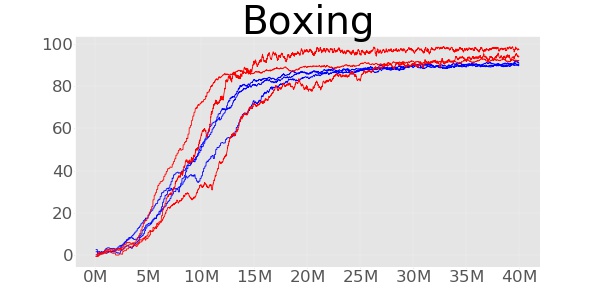}}
\hspace{0.01\linewidth}
\subfigure{
\includegraphics[width=0.2\linewidth]{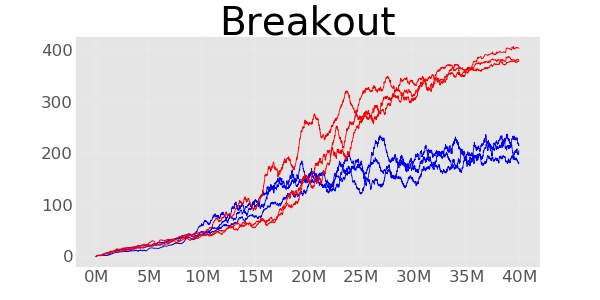}}

\subfigure{
\includegraphics[width=0.2\linewidth]{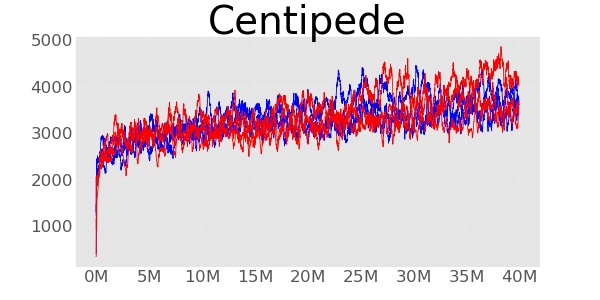}}
\hspace{0.01\linewidth}
\subfigure{
\includegraphics[width=0.2\linewidth]{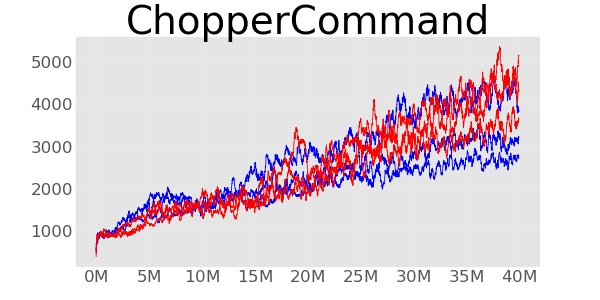}}
\hspace{0.01\linewidth}
\subfigure{
\includegraphics[width=0.2\linewidth]{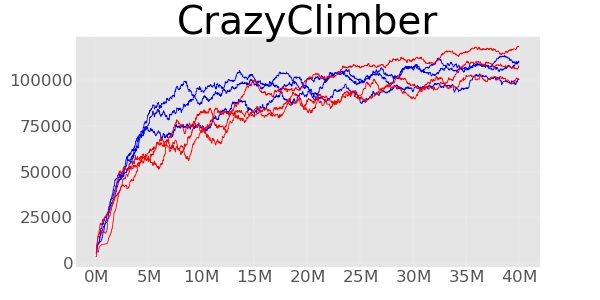}}
\hspace{0.01\linewidth}
\subfigure{
\includegraphics[width=0.2\linewidth]{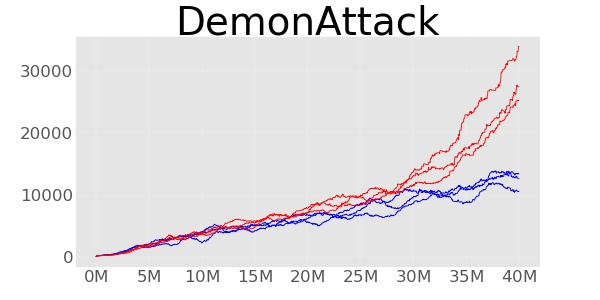}}

\subfigure{
\includegraphics[width=0.2\linewidth]{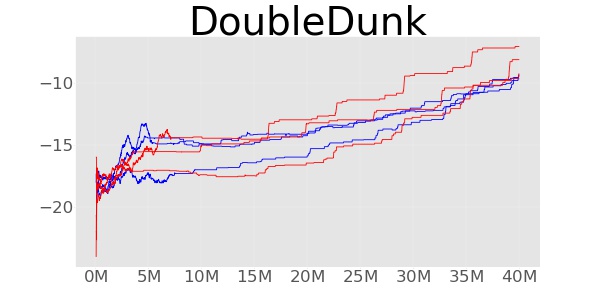}}
\hspace{0.01\linewidth}
\subfigure{
\includegraphics[width=0.2\linewidth]{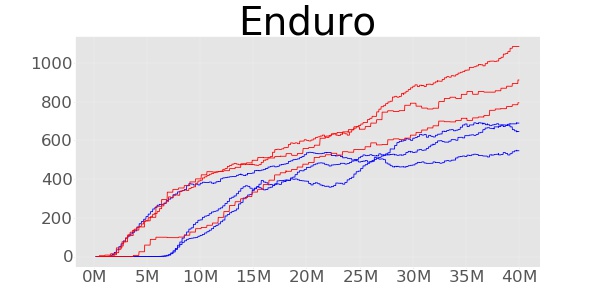}}
\hspace{0.01\linewidth}
\subfigure{
\includegraphics[width=0.2\linewidth]{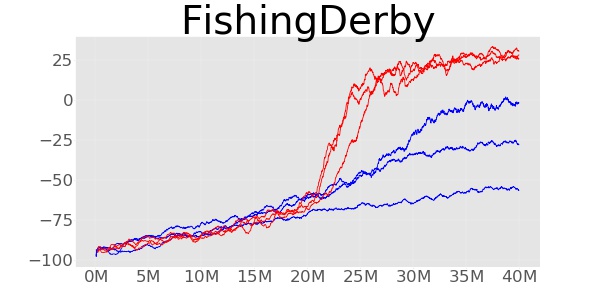}}
\hspace{0.01\linewidth}
\subfigure{
\includegraphics[width=0.2\linewidth]{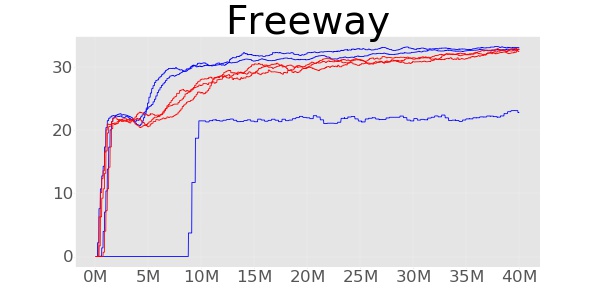}}

\subfigure{
\includegraphics[width=0.2\linewidth]{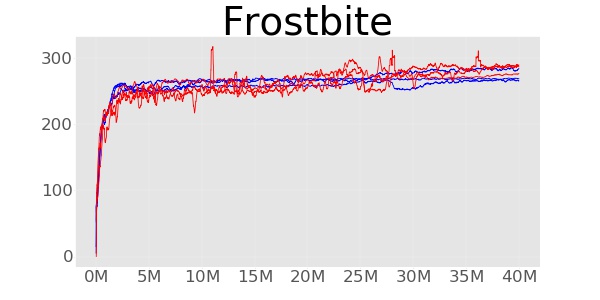}}
\hspace{0.01\linewidth}
\subfigure{
\includegraphics[width=0.2\linewidth]{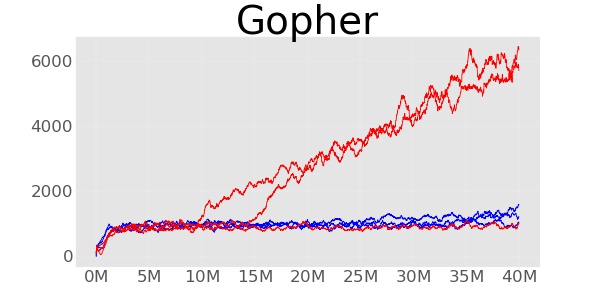}}
\hspace{0.01\linewidth}
\subfigure{
\includegraphics[width=0.2\linewidth]{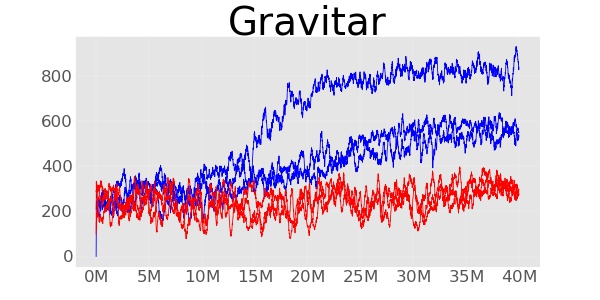}}
\hspace{0.01\linewidth}
\subfigure{
\includegraphics[width=0.2\linewidth]{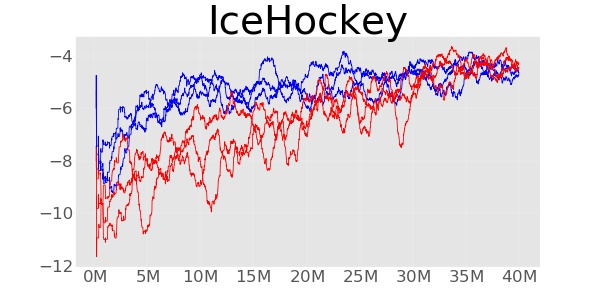}}

\subfigure{
\includegraphics[width=0.2\linewidth]{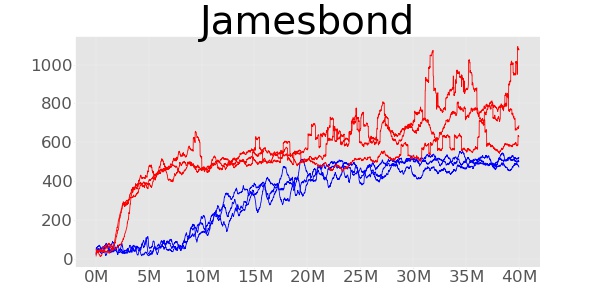}}
\hspace{0.01\linewidth}
\subfigure{
\includegraphics[width=0.2\linewidth]{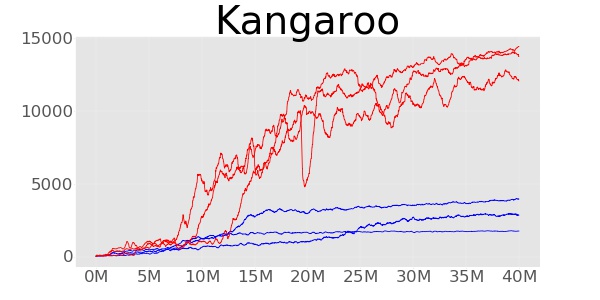}}
\hspace{0.01\linewidth}
\subfigure{
\includegraphics[width=0.2\linewidth]{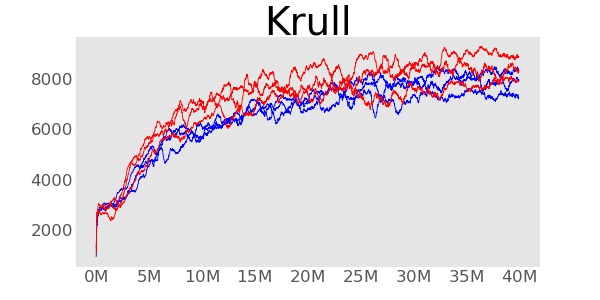}}
\hspace{0.01\linewidth}
\subfigure{
\includegraphics[width=0.2\linewidth]{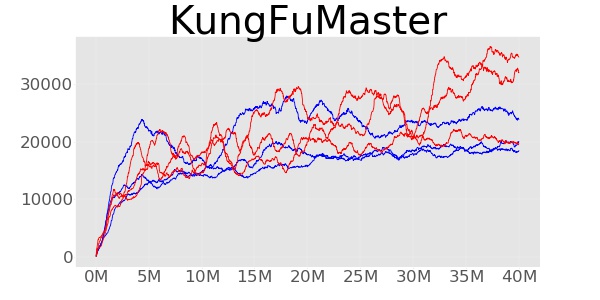}}

\subfigure{
\includegraphics[width=0.2\linewidth]{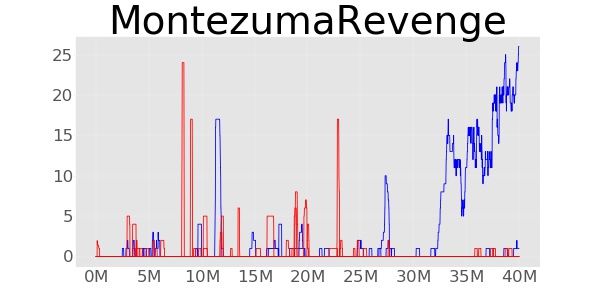}}
\hspace{0.01\linewidth}
\subfigure{
\includegraphics[width=0.2\linewidth]{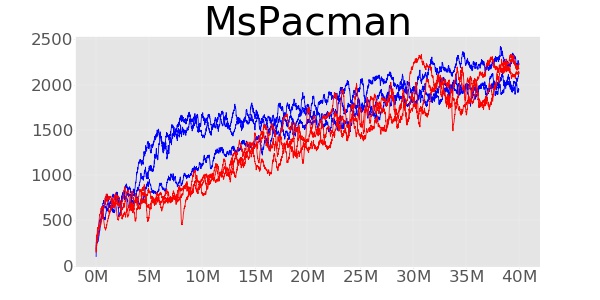}}
\hspace{0.01\linewidth}
\subfigure{
\includegraphics[width=0.2\linewidth]{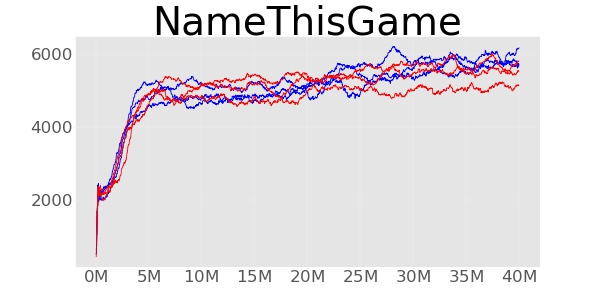}}
\hspace{0.01\linewidth}
\subfigure{
\includegraphics[width=0.2\linewidth]{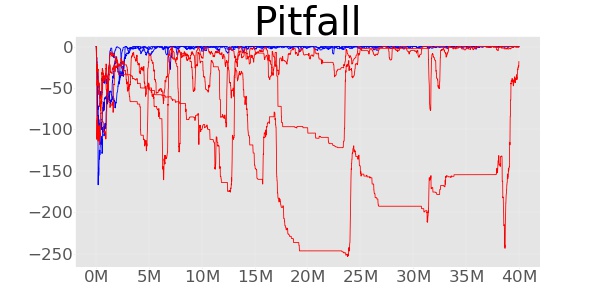}}

\subfigure{
\includegraphics[width=0.2\linewidth]{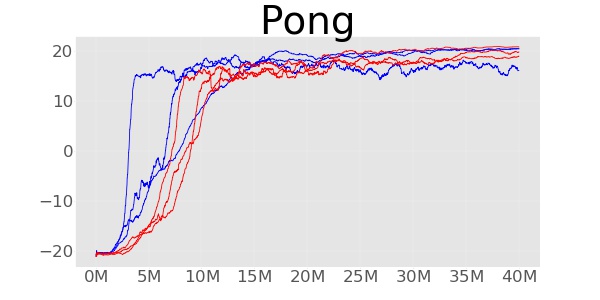}}
\hspace{0.01\linewidth}
\subfigure{
\includegraphics[width=0.2\linewidth]{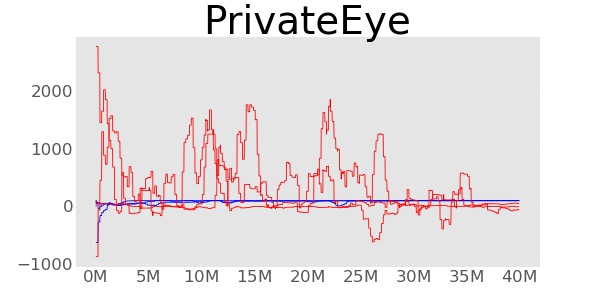}}
\hspace{0.01\linewidth}
\subfigure{
\includegraphics[width=0.2\linewidth]{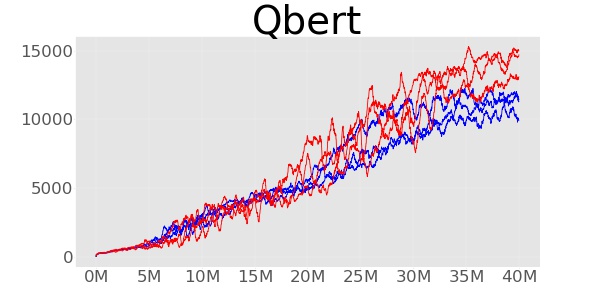}}
\hspace{0.01\linewidth}
\subfigure{
\includegraphics[width=0.2\linewidth]{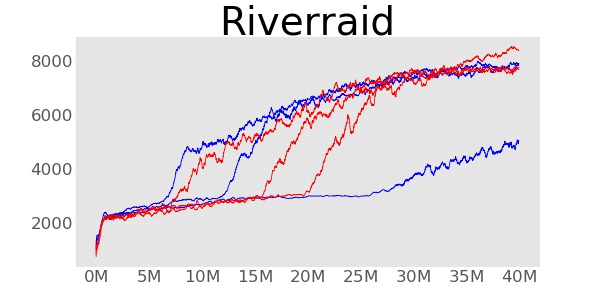}}

\subfigure{
\includegraphics[width=0.2\linewidth]{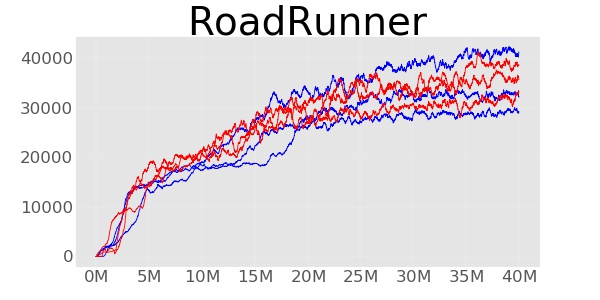}}
\hspace{0.01\linewidth}
\subfigure{
\includegraphics[width=0.2\linewidth]{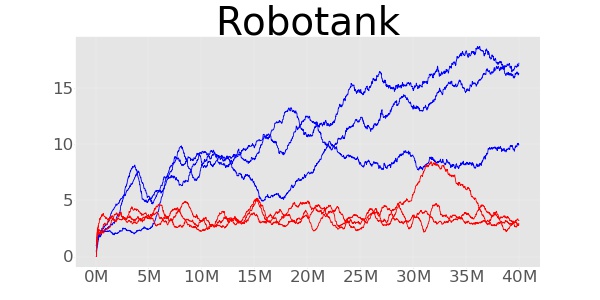}}
\hspace{0.01\linewidth}
\subfigure{
\includegraphics[width=0.2\linewidth]{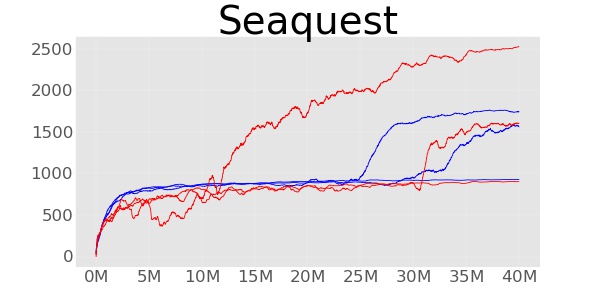}}
\hspace{0.01\linewidth}
\subfigure{
\includegraphics[width=0.2\linewidth]{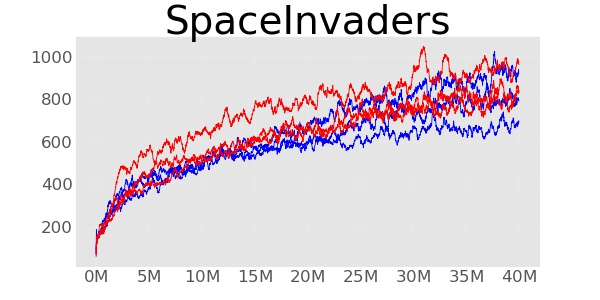}}

\subfigure{
\includegraphics[width=0.2\linewidth]{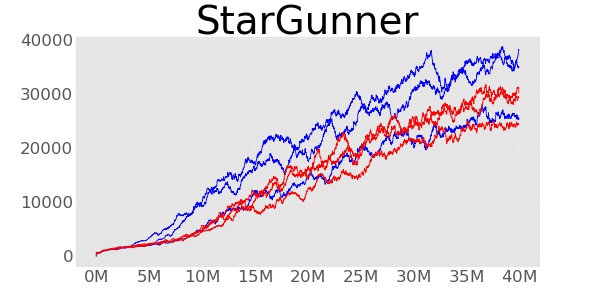}}
\hspace{0.01\linewidth}
\subfigure{
\includegraphics[width=0.2\linewidth]{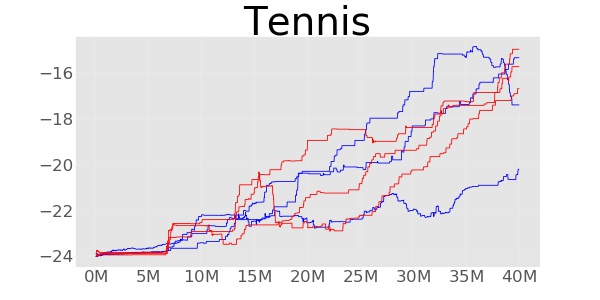}}
\hspace{0.01\linewidth}
\subfigure{
\includegraphics[width=0.2\linewidth]{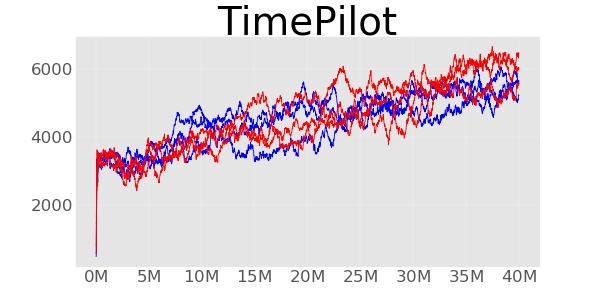}}
\hspace{0.01\linewidth}
\subfigure{
\includegraphics[width=0.2\linewidth]{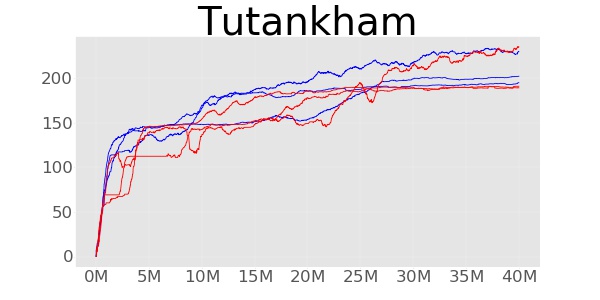}}

\subfigure{
\includegraphics[width=0.2\linewidth]{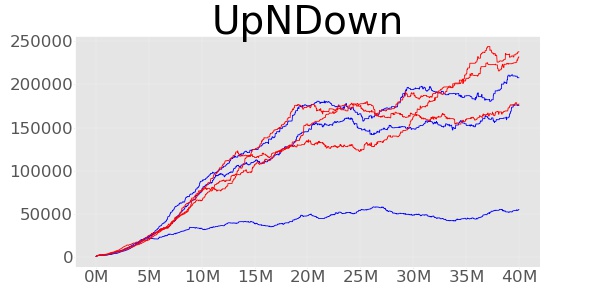}}
\subfigure{
\includegraphics[width=0.2\linewidth]{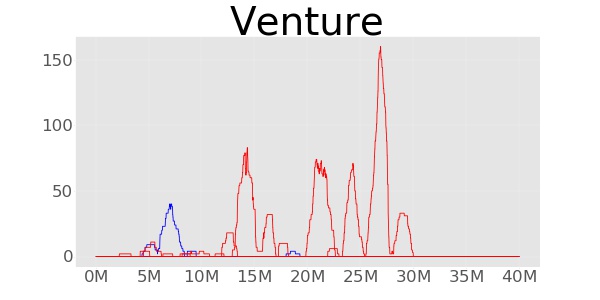}}
\hspace{0.01\linewidth}
\subfigure{
\includegraphics[width=0.2\linewidth]{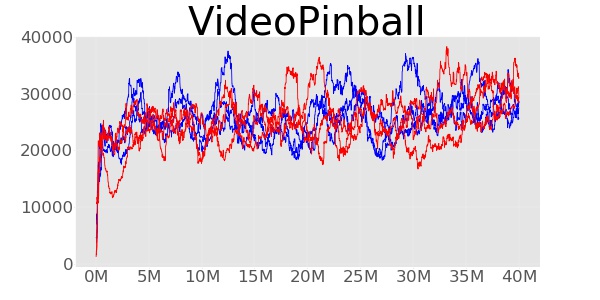}}
\hspace{0.01\linewidth}
\subfigure{
\includegraphics[width=0.2\linewidth]{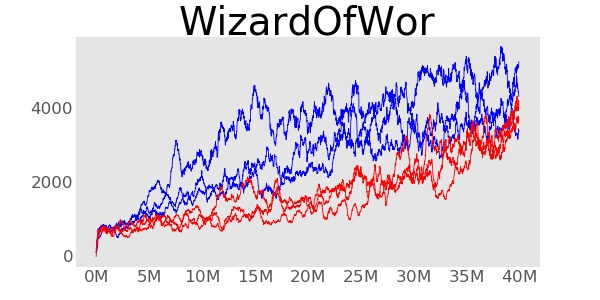}}

\subfigure{
\includegraphics[width=0.2\linewidth]{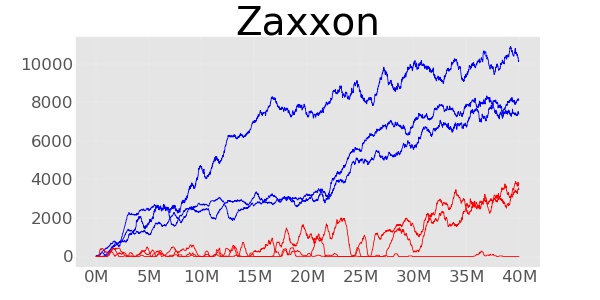}}
\hspace{0.01\linewidth}
\subfigure{
\includegraphics[width=0.2\linewidth]{tuli.jpg}}

\label{fig:subfig}
\caption{Comparison of PPO-B and PPO on all 49 ATARI games included in OpenAI Gym}
\end{figure}

\linespread{0.8}
\begin{table}[h]
\centering
\label{tab:booktabs}
\begin{tabular}{ccc}
\toprule
 & PPO-B & PPO \\
\midrule
Alien & $\pmb{1827.7}$ & 1629.73\\
Amidar & $\pmb{713.71}$ & 563.07\\
Assault & $\pmb{4154.74}$ & 3666.45 \\
Asterix & $\pmb{3017.33}$ & 2890.33 \\
Asteroids & 1276.87 & $\pmb{1867.2}$ \\
Atlantis & $\pmb{2255528.0}$ & 1956856.33 \\
BankHeist & 1160.53 & $\pmb{1192.0}$ \\
BattleZone & 4826.67 & $\pmb{18900.0}$ \\
BeamRider & 1567.07 & $\pmb{}$2809.02 \\
Bowling & $\pmb{50.91 }$& 41.06 \\
Boxing & $\pmb{94.56}$ & 90.36 \\
Breakout & $\pmb{386.8}$6 & 200.26 \\
Centipede & \pmb{3710.1} & $3604.36$ \\
ChopperCommand & $\pmb{4394.67}$ & 3280.67 \\
CrazyClimber & $\pmb{109102.0}$ & 106883.67 \\
DemonAttack & $\pmb{28893.82}$ & 12197.92 \\
DoubleDunk & $\pmb{-8.19}$ & -9.53 \\
Enduro & $\pmb{931.14}$ & 627.78 \\
FishingDerby & $\pmb{28.15}$ & -28.78 \\
Freeway & $\pmb{32.75}$ & 29.59 \\
Frostbite & $\pmb{284.6}$ & 273.1 \\
Gopher & $\pmb{4372.53}$ & 1296.47 \\
Gravitar & 282.0 & $\pmb{639.0}$ \\
IceHockey & $\pmb{-4.4}$ & -4.68 \\
Jamesbond & $\pmb{798.0}$ & 476.12 \\
Kangaroo & $\pmb{13430.33}$ & 2849.33 \\
Krull & $\pmb{8370.36}$ & 7823.15 \\
KungFuMaster & $\pmb{28758.0}$ & 20762.67 \\
MontezumaRevenge & 0.0 & $\pmb{9.0}$ \\
MsPacman & $\pmb{2181.1}$ & 2104.13 \\
NameThisGame & $\pmb{5484.43}$ & 5842.5 \\
Pitfall & -6.41 & $\pmb{0.0}$ \\
Pong & $\pmb{19.88}$ & 19.03 \\
PrivateEye & -2.45 & $\pmb{100.0}$ \\
Qbert & $\pmb{14254.25}$ & 10885.08 \\
Riverraid & $\pmb{7946.57}$ & 6921.53 \\
RoadRunner & $\pmb{35818.67}$ & 34237.0 \\
Robotank & 2.99 & $\pmb{14.44}$ \\
Seaquest & $\pmb{1677.07}$ & 1409.27 \\
SpaceInvaders & $\pmb{880.37}$ & 805.98 \\
StarGunner & 28097.33 & $\pmb{33005.33}$ \\
Tennis & $\pmb{-15.78 }$& -17.63 \\
TimePilot & $\pmb{6033.0 }$& 5474.67 \\
Tutankham & 205.35 & $\pmb{209.39}$ \\
UpNDown & $\pmb{215783.57}$ & 145988.73 \\
Venture & 0.0 & 0.0 \\
VideoPinball & $\pmb{30142.86}$ & 27183.01 \\
WizardOfWor & 3987.67 & $\pmb{4295.67}$ \\
Zaxxon & 2420.67 & $\pmb{8553.33}$ \\
\bottomrule
\end{tabular}
\caption{Mean final scores (last 100 episodes) of PPO-B and PPO on Atari games}
\end{table}

\end{document}